\documentclass[11pt]{article}

\usepackage{fullpage}

\usepackage{amsmath, amssymb}

\usepackage{doi}

\usepackage{graphicx}

\usepackage{verbatim}

\usepackage[ruled,linesnumbered,vlined]{algorithm2e}

\usepackage{multirow}

\usepackage{xspace} 
\usepackage{subfig}\DeclareCaptionType{copyrightbox}
\newcommand{\Schwefel}{\textsc{Schwefel}\xspace}
\newcommand{\Rosenbrock}{\textsc{Rosenbrock}\xspace}
\newcommand{\Ackley}{\textsc{Ackley}\xspace}
\newcommand{\Griewank}{\textsc{Griewank}\xspace}
\newcommand{\Elliptic}{H. C. \textsc{Elliptic}\xspace}
\newcommand{\Rastrigin}{\textsc{Rastrigin}\xspace}
\newcommand{\Sphere}{\textsc{Sphere}\xspace}

\usepackage{tabularx, booktabs}

\usepackage{wrapfig}
\newcolumntype{Y}{>{\centering\arraybackslash}X}

\graphicspath{{./figs/}}

\begin{document}

\title{\LARGE\bf
Towards a Better Understanding of the Local Attractor in
Particle Swarm Optimization: Speed and Solution Quality}

\author{Vanessa Lange \qquad Manuel Schmitt \qquad Rolf Wanka\\[2mm]
Department of Computer Science\\
University of Erlangen-Nuremberg, Germany\\
{\tt vanessa.lange@fau.de, \{manuel.schmitt, rolf.wanka\}@cs.fau.de}
}

\date{ }

\maketitle

\begin{abstract}
Particle Swarm Optimization (PSO) is a popular nature-in\-spired meta-heuristic for solving continuous optimization problems. Although this technique is widely used, the understanding of the mechanisms that make swarms so successful is still limited. We present the first substantial experimental investigation of the influence of the local attractor on the quality of exploration and exploitation. We compare in detail classical PSO with the social-only variant where local attractors are ignored. To measure the exploration capabilities, we determine how frequently both variants return results in the neighborhood of the global optimum. 
We measure the quality of exploitation by considering only function values from runs that reached a search point sufficiently close to the global optimum and then comparing in how many digits such values still deviate from the global minimum value. 
It turns out that the local attractor significantly improves the exploration, but sometimes reduces the quality of the exploitation. As a compromise, we propose and evaluate a hybrid PSO which switches off its local attractors 
at a certain point in time. 
The effects mentioned can also be observed by measuring the potential of the swarm.
\end{abstract}


\section{Introduction}
\label{sec:intro}

\noindent\textbf{The Particle Swarm Optimization Algorithm.}
Particle Swarm Optimization (PSO) is a popular metaheuristic designed for
solving optimization problems on continuous domains. It has been introduced by Kennedy and Eberhart~\cite{ken_eb_1995,eb_ken_1995}.
In contrast to evolutionary algorithms, the
particles of a swarm cooperate and share information about the search space
rather than competing against each other.
PSO has been applied successfully to a wide range of
optimization problems, e.\,g. in Biomedical Image Processing~\cite{WSZZE:04},
Geosciences~\cite{OD:10}, Mechanical Engineering~\cite{GWHK:09}, and
Materials Science~\cite{RPPN:09}. 
The popularity of the PSO framework in these scientific communities
is due to the fact that it on the one hand can be realized and,
if necessary, adapted to further needs easily,
but on the other hand empirically shows good performance results with respect
to the quality of the 
solution found and the speed needed to obtain it.
By adapting its parameters, users may in real-world applications
easily and successfully control the swarm's
behavior with respect to
``exploration'' (``searching where no one has searched before'')
and
``exploitation'' (``searching around a good position'').
A thorough discussion of PSO can be found in~\cite{swarmhandbook:11}.
To be precise, let a fitness function (also called objective function) $f:\mathbb{R}^D\rightarrow \mathbb{R}$
on a $D$-dimensional domain be given
that (w.\,l.\,o.\,g.) has to be minimized.
A population of {\it particles}, each consisting of a \emph{position} (the candidate for
a solution), a \emph{velocity}
and a \emph{local attractor} (also referred to as \emph{private guide}), moves through the search space $\mathbb{R}^D$.
The local attractor of a particle is the best position with respect to $f$
this particle has encountered so far. Additionally, the swarm has a common
memory, the \emph{global attractor} (also referred to as \emph{local guide}), which is the best position any particle has found
so far.
The movement is governed by the so-called \emph{movement equations}.

Many variants of PSO have been developed and empirically proven to be efficient.
Most of them extend the classical PSO algorithm by additional operations.
As just one example out of many, van den Bergh/Engelbrecht~\cite{BE:02} substantially modify
the movement equations, enabling the particles to count the number of times they
improve the global attractor and use this information.

Although the efficiency of PSO is widely known, the understanding of 
the mechanisms that make the swarm so successful is still limited. A theoretical
analysis of the particles' trajectories can be found in \cite{CK:02}. Parameter
selection guidelines guaranteeing the convergence of the swarm under the assumption
of fixed attractors have been developed in \cite{JLY:07a}. Additional guidelines,
for which the classical swarm in the $1$-dimensional case and a slightly modified
PSO in the $D$-dimensional case finds provably at least a local optimum of any sufficiently
smooth function can be found in \cite{SW:13}.
In~\cite{SW:13}, also the notion of the \emph{potential}
of a particle swarm has been introduced.
The potential, which we will use also in this paper,
is a measure for the swarm's capability to reach search points far away
from the current global attractor.

In this paper, we investigate the influence of the local 
attractor on the speed of convergence and the quality of the found solution.

In \cite{LW:11}, the authors prove that a swarm consisting of only a single particle
does, with positive probability, not converge towards a local optimum. Therefore, the
importance 
of the global attractor is beyond doubt since without it, the swarm would
act like many completely independent $1$-particle swarms.
But to the best of our knowledge, 
the exact influence 
of the local attractor has not yet been formally addressed. 
Closest to that direction, Kennedy~\cite{K:97} applied, among other simplified versions
of PSO, the so-called ``social-only model,'' which consists of a particle swarm
without local attractors, to an artificial neuronal network learning problem. He
already noticed a ``slight susceptibility to be captured by local optima.''
Pedersen and Chipperfield~\cite{PC:10} proposed a Meta-Optimizer for finding good parameter
choices of both classical PSO and again the variant with disabled local attractors, which
they call Many Optimizing Liaisons. However, the question of the benefit from using the
local attractor remains unsolved. That is why this paper is dedicated to the particles'
local attractors.

\smallskip

\noindent\textbf{Our contribution.}
\nocite{JLY:07}
Our main goal is to better understand the influence of the local attractor on the swarm's behavior.
In order to measure the benefit 
of the local attractor, we compare 
classical PSO
with social-only PSO where we 
ignore the local attractors. 
We explain, why the local attractor is important for exploration and
for improving 
the chances of
finding not only an arbitrary local optimum, but often the global optimum of the fitness function  (it helps to leave `traps' of local optima).
Additionally, we give empirical evidence that the influence of the local attractors is significant.
However, sometimes the price of this improved exploration is a delay in the convergence. 
Here, we only consider those runs of both variants that actually (after some fixed time has been expired)
come close to the global optimum and compare which results come closer to this optimum.
The experiments show that on some fitness functions and with fixed time budget,
classical PSO performs worse than social-only PSO.
That means: The local attractor directs `in general' the swarm to better regions,
and without the local attractor the swarm finds better solutions in such a region.
Hence, we propose 
a hybrid method, applying the swarm with the
local attractors for the first half of the iterations and disabling them afterwards. Indeed, both the
quality 
of the obtained solution and the frequency of runs finding the global optimum are in between the
respective values of the classical and the social-only PSO. Finally, measuring the course of the potential of the
hybrid algorithm illustrates the increase in convergence speed by switching off the local attractors.

The paper is organized as follows: In Section \ref{sec:prelim}, we state the relevant definitions of the
PSO algorithm and the notion of potential we use for our experiments. In Section \ref{sec:setting}, the general 
setting of the performed experiments is described. In Section \ref{sec:results_modified}, the results of the
comparison between the classical and the social-only PSO are presented. Finally, in Section \ref{sec:hybrid}, one 
can find our results regarding the hybrid PSO.


\section{Preliminaries}\label{sec:prelim}

Let $D$ be the dimension of the fitness function $f$ that should be (w.\,l.\,o.\,g.) minimized. 
A particle swarm consists of $N$ particles. At every time, each particle~$i$ has a position $\vec x_i\in\mathbb{R}^D$, representing a point in the search space (being a possible solution) and a velocity $\vec v_i\in\mathbb{R}^D$. Additionally, particle $i$ has a local attractor $\vec p_i\in\mathbb{R}^D$, the best point it has visited so far. Finally, the swarm shares the global attractor $\vec p_{\rm glob}\in\mathbb{R}^D$, the best point \emph{any} particle has visited so far. Algorithm \ref{alg:clPSO} provides an overview over the classical PSO algorithm. In particular, the movement of the swarm is governed by the so called \emph{movement equations} in lines \ref{vup} and \ref{xup}. Here, $\odot$ denotes
entrywise multiplication (Hadamard product), and $\vec{r}_{\rm glob}$ and $\vec{r}_{\rm loc}$ are random vectors with entries chosen u.\,a.\,r. from $[0,1]$ at each occurrence. Moreover, $a$, $b_{\rm glob}$ and $b_{\rm loc}$ are constant weights, the so-called \emph{PSO parameters}. As recommended in \cite{CK:02}, we choose $a=0.72984$ and $b_{\rm loc}=b_{\rm glob}=1.496172$.

\begin{algorithm}[tbh]
\caption{\label{alg:clPSO}Classical PSO}

\SetKwInOut{Input}{input}
\SetKwInOut{Output}{output}

\Input{$f:\mathbb{R}^D\to\mathbb{R}$, number $N$ of particles, maxiter}
\Output{$\vec{p}_{\rm glob}\in\mathbb{R}^D$}
\For{$i=1 \to N$}{
  Initialize $\vec x_i$ randomly; 
  \enspace Initialize $\vec v_i := 0$;
  \enspace Initialize $\vec p_i := \vec x_i$\;
  }
Initialize $\vec p_{\rm glob}:=\operatorname{argmin}\{f(\vec p_i)\mid i=1\ldots N\}$\;
\For{$k=1 \to {\rm maxiter}$}{
  \For{$i=1 \to N$}{
	 $\vec{v}_{i} := a \cdot \vec{v}_{i} + b_{\rm glob} \cdot  \vec{r}_{\rm glob} \odot (\vec{p}_{\rm glob} - \vec{x}_{i}) 
         + b_{\rm loc} \cdot  \vec{r}_{\rm loc} \odot (\vec{p}_i - \vec{x}_{i})$\label{vup}\;
      $\vec{x}_i := \vec{x}_i+\vec{v}_i$\label{xup}\;
	}
  \For{$i=1 \to N$}{
    \lIf{$f(\vec{x}_i)\le f(\vec{p}_i)$}{
      $\vec{p}_i := \vec{x}_i$\;
    }
    \lIf{$f(\vec{x}_i)\le f(\vec{p}_{\rm glob})$}{
      $\vec{p}_{\rm glob} := \vec{x}_i$\;
    }  
  }
}

\end{algorithm}

Since we are interested in the benefit the local attractor has 
for the algorithm, we also study a version that has been called the social-only PSO (\cite{K:97}) and is obtained by setting $b_{\rm loc}=0$ and therefore making the particles ignore their personal memory. For a fair comparison, since a lot of effort has been put into finding good parameters for the classical PSO, we tested several parameter settings for the social-only PSO. 
Experiments have shown that 
using $a=0.72984$, $b_{\rm glob}=1.496172$ is continuing to be reasonable.

For our investigations, we will also use the notion of the so-called potential $\vec \Phi$ of the swarm as a measure for its movement in the different dimensions $d\in\{1,\ldots, D\}$ as introduced in \cite{SW:13}. Since there are different ways to measure the potential, we use the following formulation:
$$
(\vec \Phi)^d := \sqrt{\sum_{i=1}^N 2.5\cdot |(\vec v_i)^d| + |(\vec p_{\rm glob})^d-(\vec x_i)^d|}
\enspace,
$$
where $(.)^d$ means the $d$th entry of the vector.
Additionally, to measure the potential of 
a single particle $i$ instead of the potential of the complete swarm, we use
$$
(\vec \Psi_i)^d := a \cdot|(\vec v_i)^d| + b_{\rm glob} \cdot|(\vec p_{\rm glob})^d-(\vec x_i)^d|+ b_{\rm loc} \cdot|(\vec p_i)^d-(\vec x_i)^d|.
$$


\section{Setting}
\label{sec:setting}

Our experiments were performed with the following setting:
\begin{itemize}
\item The swarm size was set to $N=100$, the number of iterations
to ${\rm maxiter}=500$. 
\item We investigated the functions \Ackley, \Griewank,
 \Rastrigin, \Rosenbrock, \Schwefel, \Sphere
(for a description of these functions, see, e.\,g.,~\cite[p.~94ff]{Helwig10}), and the non-shifted, non-rotated High Conditioned \textsc{Elliptic}~\cite{CEC05Bench}.
\item For all functions, we tested all dimensionalities $D\in\{1,2,3,4,5, 10\}$.
\item Since every considered function has a bounded search space $\cal I$, we used the bound handling method \emph{Random} (\cite{ZXB:04}), i.\,e.,
if in dimension $d$ a particle leaves the search space,
the $d$th entry of its position is set randomly to a value inside the search boundaries. 
\item The particles' postitons were initialized u.\,a.\,r. over $\cal I$, the velocities were initialized with $0$.
\item Every run of Algorithm~\ref{alg:clPSO}
was repeated $50$ times. 
\end{itemize}
Note that the actual global optimum of every of the considered fitness functions was at $0$, where the double numbers have the highest precision. The reasen for the comparatively high swarm size for the rather low search space dimensions is that finding the global optimum for highly multi-modal functions is difficult. For smaller swarm sizes or higher search space dimensions, none of the studied PSO variants would have found the global optimum, preventing any meaningful comparison.
We used the following criteria to classify the obtained solution on the benchmark functions as a `local optimum' or even the `global optimum.' Note that the exact values of the global optima are known.

\smallskip
\noindent
\textbf{Global Optimum (G).}
For all functions except for \Schwefel and \Rosenbrock, we categorize a result as `global optimum' if each dimension differs by at most $0.0015 \cdot |{\cal I}|$ from the position of the global minimum. This value guarantees that no other local optimum than the global optimum itself is detected as the global minimum even for the highly multi-modal functions.

For \Schwefel and \Rosenbrock, the value was set to $0.005 \cdot |{\cal I}|$ instead, since with the value above too many results were falsely classified `O' (see below). 

\smallskip

\noindent
\textbf{Local Optimum (L).}
A result is classified as a `local minimum' if it is not classified as the global optimum and the absolute value of the derivative of the function is $\le 0.1$ in each dimension. Tests showed that only the low-dimensional ($D\le 3$) \Rosenbrock function has regions flat enough to lead to a wrong classification. Therefore, here the classification explicitly uses the fact that \Rosenbrock has only one local optimum for $D\le 3$ (\cite{Shang:2006:NER:1118006.1118014}).

\smallskip

\noindent
\textbf{Otherwise (O).}
The obtained solution is classified `O' otherwise because it is still far away from the global and any local optimum.

\medskip

This classification serves as a measure for the exploration capability of the swarm, i.\,e., the better the swarm explores, the more results should be classified as (G). Additionally to the classification, 
we collected for each fitness function $f$ all fitness values of the results that were classified as global optimum and calculated their average in order to measure the quality of the exploitation on $f$.  
The obtained value will be referred to as \emph{precision}.


\section{Results of the Social-Only PSO Algorithm}
\label{sec:results_modified}

We examined our results under two different aspects. First, we focused on the exploration behavior of the PSO and measured, how frequently the obtained result could be classified as the global optimum. Afterwards, we studied the exploitation capabilities by comparing how close the results that were categorized as global optimum came to the actual global minimum.

\subsection{Impact of the Local Attractor on Exploration}
\label{subsec:modified_optima}
We wanted to examine the influence of the local attractor on the capability of the PSO to converge towards the global optimum of our benchmark functions. 

Table \ref{tab:socialPSO_optima} shows the results for $50$ runs with $D=3$ and, for reasons to be stated later in this section, also for the $4$-dimensional \Rastrigin. As one can see, with respect to the behavior of the PSOs, two essentially different classes among the fitness functions can be distinguished.

\begin{table}
\caption{Comparison of the classification results from the classical and the social-only PSO processing various $3$-dimensional and a $4$-dimensional function. For \Griewank, \Rastrigin and \Schwefel, classical PSO shows a significantly better exploration.}
\label{tab:socialPSO_optima}
\center{
\def\arraystretch{1.2}

\begin{tabularx}{0.7\textwidth}{|c||Y|Y|Y||Y|Y|Y|}

\hline

 \multicolumn{1}{ |c|| }{}
 & \multicolumn{3}{ c|| }{Classical PSO}
 & \multicolumn{3}{ c| }{Social-only PSO} \\

\hline 

 Function & G & L & O & G & L & O \\
\hline \hline

\Ackley & 50 & 0 & 0 & 50 & 0 & 0 \\
\Griewank & {\bf 25} & 25 & 0 & 2 & 48 & 0 \\ 
\Elliptic & 50 & 0 & 0 & 50 & 0 & 0 \\
\Rastrigin (3-dim.) & {\bf 50} & 0 & 0 & 28 & 22 & 0 \\
\Rosenbrock & 47 & 0 & 3 & 50 & 0 & 0 \\
\Schwefel & {\bf 50} & 0 & 0 & 36 & 3 & 11 \\ 
\Sphere & 50 & 0 & 0 & 50 & 0 & 0 \\
\hline \hline
\Rastrigin (4-dim.) & {\bf 50} & 0 & 0 & 10 & 40 & 0\\
\hline

\end{tabularx}

}
\end{table} 

Within the first group, consisting of \Ackley, High Conditioned \textsc{Elliptic}, \Rosenbrock and \Sphere, the global optimum was easily found. The social-only as well as the unmodified PSO algorithm brought good results, finding the global minimum in every or almost every run. Since the High Conditioned \textsc{Elliptic}, the ($3$-dimensional) \Rosenbrock and the \Sphere function are unimodal, the importance of exploring the search space and therefore of the local attractor itself is comparatively small. Although the \Ackley function is not unimodal, there are major differences in location and function value between the global optimum and the other local optima. Therefore, even the limited exploration capability of social-only PSO is still sufficient. Manual checks on the \Rosenbrock runs revealed that the results classified as `O' usually were close to the bound for being classified as global optimum, but had a precision slightly too poor. Therefore, these results are not caused by a weakness in exploration but in exploitation.

In the second group, consisting of \Griewank, \Rastrigin and \Schwefel, the outputs of social-only PSO were considerably worse. While the unmodified, i.\,e., classical PSO algorithm could
still solve the optimization problems in most cases, the social-only algorithm often failed to find the global optimum. 
As a matter of fact, these functions all have a large number of local optima around the global optimum, some of them with values close to the global minimum. Therefore, exploring the search space is vital for finding the global optimum and without the local attractors, social-only PSO gets trapped into local optima more easily. 

The \Griewank function was picked for further investigation because here the difference between the two PSO versions was already visible in the $2$-dimensional case. The function itself  
has the form 
$f(\vec x) = \mu\cdot\sum_{d=1}^D  x_d^2 -\prod_{d=1}^D\cos(x_d/\sqrt{d})+1$, 
i.\,e., it consists of \Sphere multiplied with a small factor $\mu$, set to $1/4000$ in the original problem formulation, and is covered with ``noise'' in form of cosine oscillations.
Our idea was to increase the weight $\mu$ of the sphere part in order to strengthen its influence and to increase the difference between the values of neighboring local minima. The results are presented in Table \ref{tab:griewankN_optima}. As one can see, for both PSO versions, the more \Sphere gained influence, the more reliably was the global minimum found.

\begin{table}
\caption{Comparison of the classification results for the classical PSO and the
social-only PSO processing the $5$-dimensional \Griewank function with different weights $\mu$ of its \Sphere part.}
\label{tab:griewankN_optima}
\center{
\def\arraystretch{1.2}
\begin{tabularx}{0.7\textwidth}{|c||Y|Y|Y||Y|Y|Y|}
\hline

 \multicolumn{1}{ |c|| }{}
 & \multicolumn{3}{ c|| }{Classical PSO}
 & \multicolumn{3}{ c| }{Social-only PSO} \\
\hline 

  \Griewank's $\mu$ & G & L & O & G & L & O \\
\hline \hline

$1/4000$ & 3 & 47 & 0 & 0 & 50 & 0 \\ 
$1/2000$ & 16 & 34 & 0 & 1 & 49 & 0 \\
$1/1000$ & 20 & 30 & 0 & 1 & 49 & 0 \\
$1/500$ & 34 & 16 & 0 & 7 & 43 & 0 \\
$1/100$ & 50 & 0 & 0 & 38 & 12 & 0 \\ 
$1/10$ & 50 & 0 & 0 & 50 & 0 & 0 \\
\hline

\end{tabularx}

}
\end{table} 

For an explanation of the much better success rate of unmodified PSO in finding the best among many similar local optima,
consider a function with unique global optimum $\vec x_*$ and many 
local optima with function values close to $f(\vec x_*)$ around $\vec x_*$. This is the case, e.\,g., on the \Griewank function. Since the particles are uniformly distributed over the search space, there is a certain probability for the global attractor $\vec p_{\rm glob}$ to be closer to a local optimum $\vec x_L$ than to $\vec x_*$. On the other hand, there is a good chance for some particle $i$ to have at least its local attractor $\vec p_i$ close to 

\begin{wrapfigure}{r}{0.6\textwidth}
\centering
\includegraphics[width=0.55\textwidth]{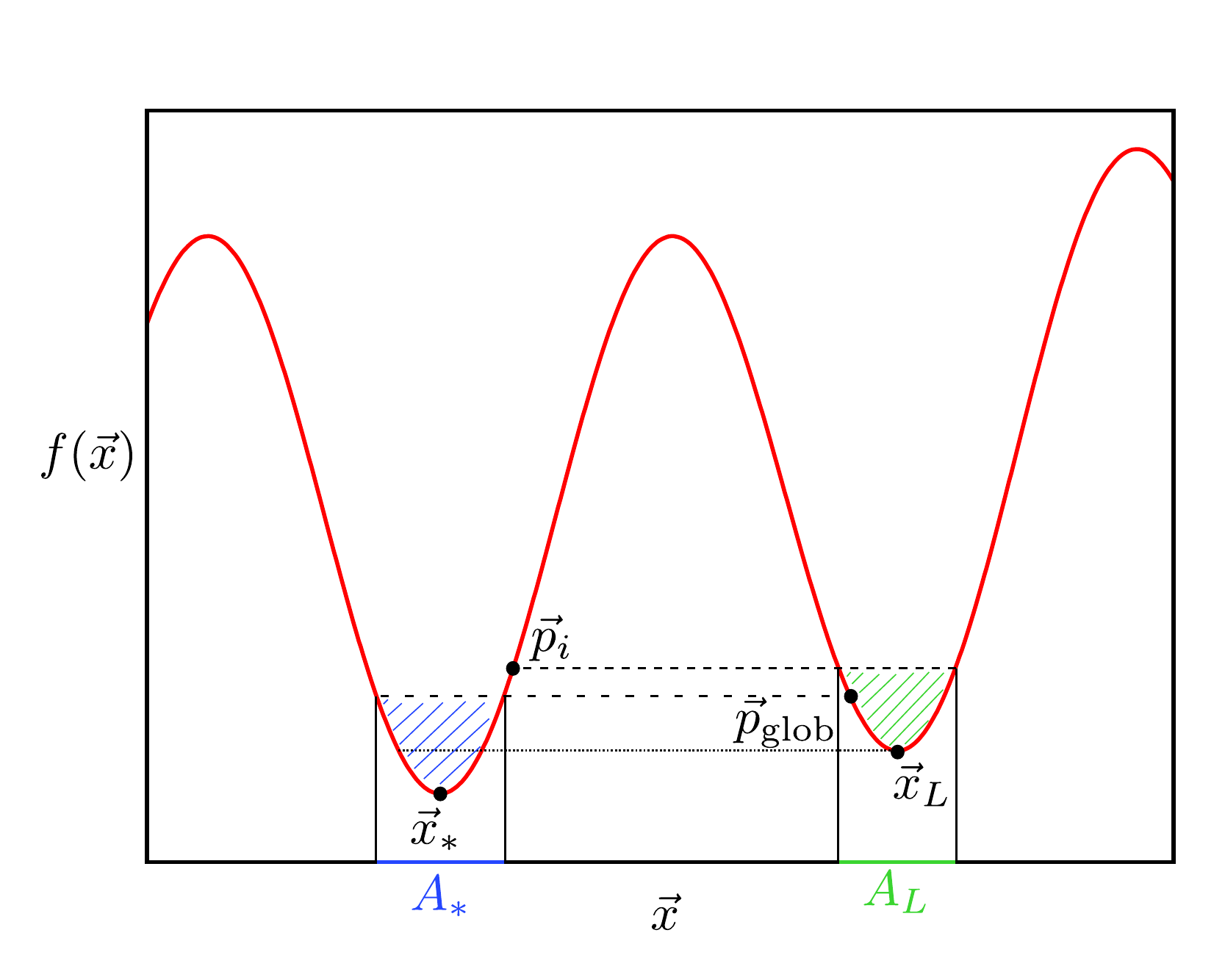}
\caption{Pathological situation of a particle with local attractor near the global optimum and global attractor near a non-global but local optimum; in order to convince the swarm for the global optimum, the particle must hit $A_*$.}
\label{fig:griewank}
\end{wrapfigure}
\noindent
$\vec x_*$. Figure \ref{fig:griewank} presents such a situation. 
While the probability for such a configuration 
to occur does not depend much on the use of the local attractor, we will see that the probability for the global attractor to enter the valley of the global optimum after the occurrence of such a situation indeed does.
Let $A_*$ denote the region around $\vec x_*$ consisting of all points better
than $\vec p_{\rm glob}$, and let $A_L$ denote the region of all points around $\vec x_L$ 
better than $\vec p_i$. One can think of $A_*$ as the region that after entering allows particle $i$ to update the global attractor and therefore convince the whole swarm to start searching in this region. Similarly, $A_L$ can be thought of as the region which upon entering makes particle $i$ update its local attractor and consequently forget about the valley of the global minimum. For 
social-only PSO, i.\,e., when the swarm is not influenced by the local attractor, for hitting $A_*$ a particle needs in every dimension $d$ potential $(\vec \Psi)^d$ of order $|(\vec x_*)^d-(\vec p_i)^d|$ to overcome the distance between $\vec p_i$ and $\vec x_*$. Consequently, even with sufficiently high potential, the probability of hitting $A_*$ is of order at most $\approx |A_*|/(|Q(\vec x_*,\vec x_L)|)$ where $Q(\vec a,\vec b) = [(\vec a)^1,(\vec b)^1]\times\ldots\times[(\vec a)^D,(\vec b)^D]$ is the smallest paraxial box containing $\vec a$ and $\vec b$. Under the assumption that meanwhile the global attractor is not altered substantially by the remaining swarm, the potential drops and after some iterations it falls below a certain bound. Then $\vec x_*$ is out of reach and particle $i$ has no chance to lure the swarm towards $\vec x_*$ anymore. 

On the other hand, if the local attractor is present, the chance of the particle hitting $A_*$ is also of order $\approx |A_*|/(|Q(\vec x_*,\vec x_L)|)$, but since the distance of both attractors maintains the necessary potential level and therefore prevents $\vec x_*$ from getting out of reach, the probability for hitting $A_*$ does not vanish until particle $i$ updates its local attractor by hitting $A_L$. The overall chance of hitting $A_*$ before hitting $A_L$ is of order $|A_*|/(|A_*|+|A_L|)$. For small values of $|A_*|$ and $|A_L|$, this success probability is considerably larger than $|A_*|/|Q(\vec x_*,\vec x_L)|$. Furthermore, even if the unmodified particle hits $A_L$ before $A_*$, then in the next iteration after the respective local attractor update, particle $i$ is likely to have still a sufficient potential to reach $x_*$ and is therefore in a situation not worse than the situation of the social-only particle.

\subsection{Impact of the Local Attractor on Exploitation}
\label{subsec:modified_precision}
After having analyzed how often the global optimum was found by the social-only PSO algorithm, we now focus on the precision of the results. 
We calculated the arithmetic mean of the function values of the PSOs' results subtracted by the known function value of the global minimum, taking only the runs into account which actually found the global optimum. The calculations were done using Java 1.7 and Python 2.7.3 which work with double precision on the chosen architecture. 
For our examination this was sufficiently precise. The results are shown in Table \ref{tab:socialPSO_precision}. 

Since both PSO versions reached the limit of double precision when processing the $3$-dimensional \Rastrigin, we added the results of the $4$-dimensional \Rastrigin to make the differences in precision visible. One can see that the precision of the results the social-only PSO algorithm returned 
was often better and never 
extremely worse than the precision of the unmodified algorithm. 
It is noticeable that sometimes (\Griewank, \Rosenbrock, \Rastrigin for at least $4$ dimensions) the precision was even significantly better. Manual checks confirmed that this significance is not an artifact of the functions' shapes but that the obtained positions were significantly closer to the optimum.
Since the presence of the local attractor improves exploration, it is natural to assume that it harms exploitation by a certain amount, so disabling the local attractors might result in a higher precision of the result. 

\begin{table}
\caption{Comparison of the precision of the classical PSO and the social-only PSO processing various $3$-dimensional and a $4$-dimensional function. For \Griewank, \Rosenbrock and \Rastrigin (4-dim.), classical PSO shows a significantly worse precision.}
\label{tab:socialPSO_precision}
\center{
\def\arraystretch{1.2}
\begin{tabularx}{0.7\textwidth}{|c| |Y|Y|}
\hline

 \multicolumn{1}{ |c|| }{}
 & \multicolumn{1}{ c| }{Classical PSO}
 & \multicolumn{1}{ c| }{Social-only PSO} \\
\hline

 Function & Precision & Precision\\
\hline \hline

\Ackley & 4.4409e-16  & 1.2967e-15  \\
\Griewank & 3.6068e-12 & {\bf 0.0} \\ 
\Elliptic & 4.0877e-37 & 8.2328e-40 \\
\Rastrigin (3-dim.) & 0.0 & 0.0 \\
\Rosenbrock & 0.0011 & {\bf 8.5184e-20} \\
\Schwefel & 0.0776 & 0.8312 \\ 
\Sphere &6.0717e-42 & 7.2010e-44 \\
\hline \hline
\Rastrigin (4-dim.) & 4.6544e-07 & {\bf 0.0}\\
\hline

\end{tabularx}
}
\end{table}

\begin{figure}
\centering
\includegraphics[width=0.55\textwidth]{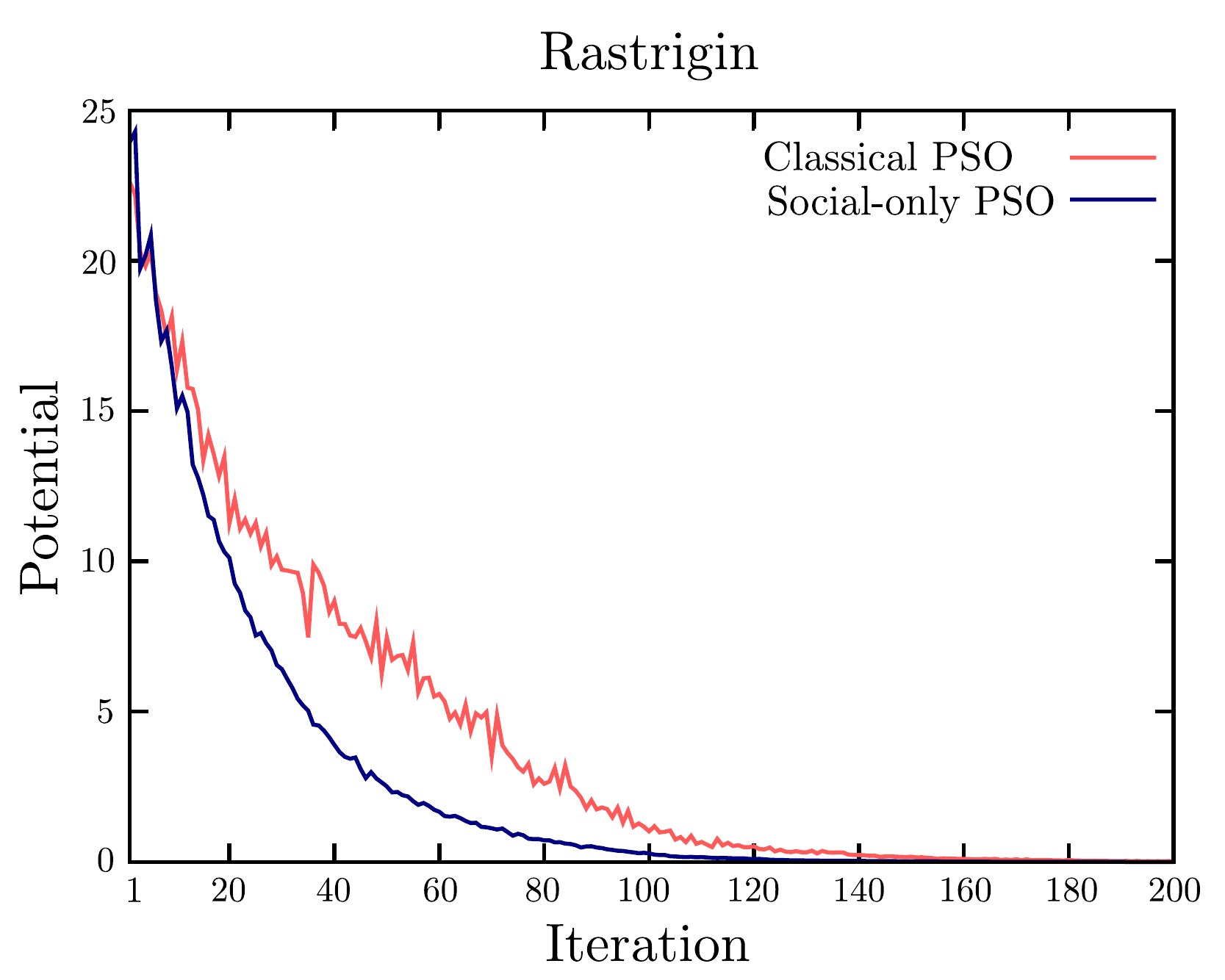}
\caption{Course of the potential of the first $200$ iterations obtained from a sample run for both the classical and the social-only PSO, processing the $1$-dimensional \Rastrigin function.} 
\label{fig:rastrigin_potential_socialOnly} 
\end{figure}

In order to further illustrate that the local attractor indeed improves exploration at the cost of exploitation and that there is indeed a measurable effect, we analyzed the potential of the swarm. 
If the potential of a swarm tends toward zero, the swarm converges (\cite{SW:13}). We measured 
the course of the potential for both algorithms over the iterations.
The obtained measurements have 
in common that the swarms of the social-only PSO algorithm 
lost their potential faster. The difference to the unmodified swarm was sometimes very close, but in many cases 
clearly visible or even considerably big. For example, Figure \ref{fig:rastrigin_potential_socialOnly} 
shows the course of the potential obtained from a sample run of both
classical and social-only PSO, processing $1$-dimensional \Rastrigin. 

\section{Hybrid Approach}
\label{sec:hybrid}

According to the results in the previous section, the social-only PSO algorithm often
(as seen in Section~\ref{subsec:modified_optima}) fails
in finding the global optimum of strongly multi-modal functions. On the other hand,
we have empirical evidence from Section~\ref{subsec:modified_precision}
that for certain settings the precision of the output is
better in comparison to the precision of the output found by the classical PSO algorithm. 
Due to these observations 
the idea is 
to build a hybrid PSO algorithm that preferably balances out the disadvantages and can still rely on the mentioned advantages.

\subsection{Development of the Hybrid PSO Algorithm}
We can assume that in a typical run, at a certain stage the swarm has made its definitive choice for one local optimum and spends the remaining time on exploitation in order to increase the precision. From that point on, the local attractor appears to be a drawback. Therefore, in our hybrid approach, after, say, half of the iterations we 
switch $b_{\rm loc}$ from its initial value to $0$ to increase the convergence rate of the swarm.
Note that this is a somewhat arbitrary choice made to further emphasize the influence of the local attractor. For a competetive new PSO variant, future work will have to include more refined parameter tunings. Other combinations of social-only PSO and classical PSO are also possible, e.\,g., one could use a heterogenous swarm consisting of particles that utilize their local attractors and others that do not.

\subsection{Results of the Hybrid PSO Algorithm}
\begin{figure}[tb] 
\centering
\subfloat[Overview over the complete run.]{
  \includegraphics[width=0.4\textwidth]{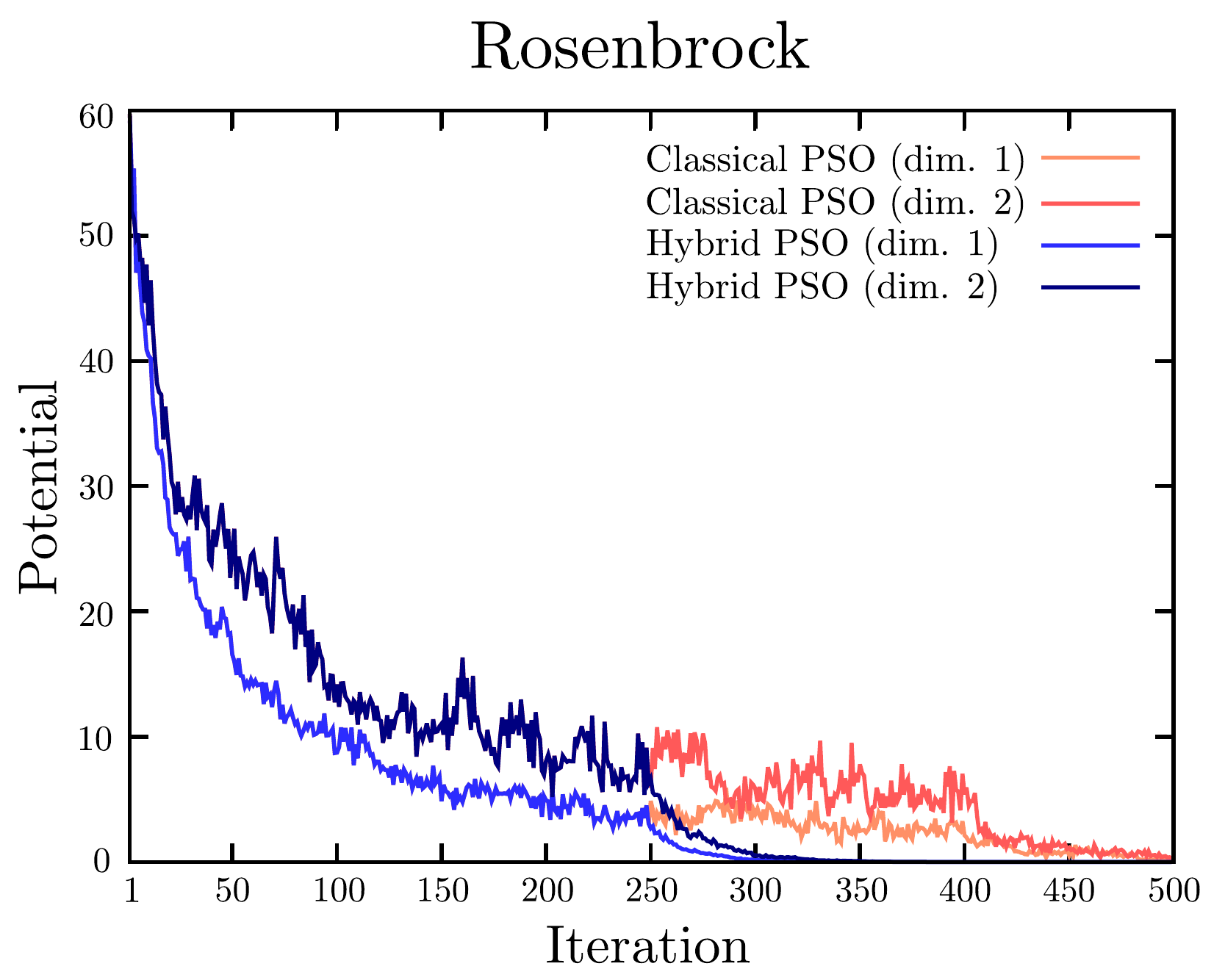}
}\qquad\qquad\qquad
\subfloat[Magnified extract between iteration $200$ and $350$.]{
  \includegraphics[width=0.4\textwidth]{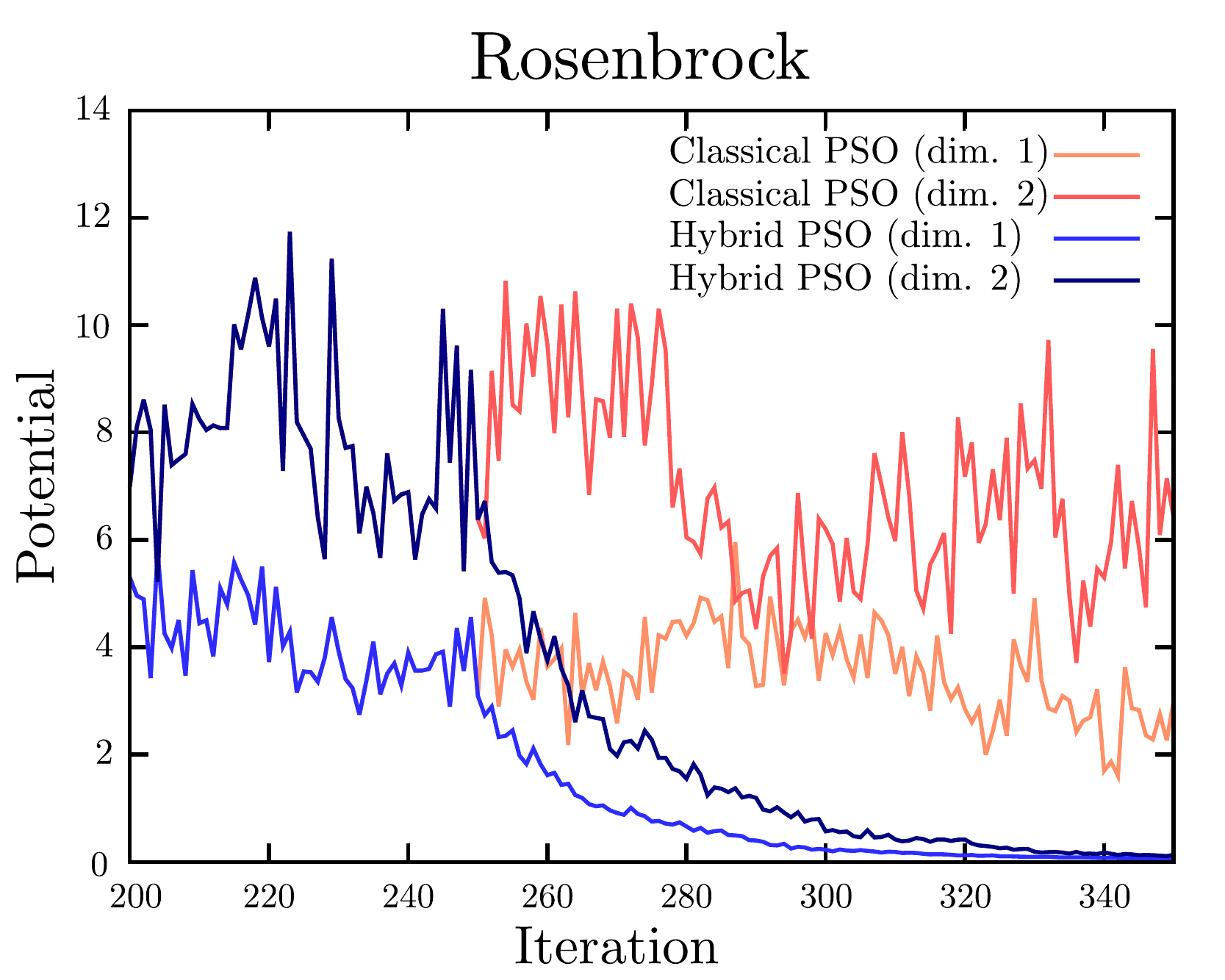}
}
\caption{Course of the potentials from a sample run for both the classical and the hybrid PSO processing the $2$-dimensional \Rosenbrock function.}
\label{fig:rosenbrock_potential_hybrid}
\end{figure}
The test results once again document 
our assumptions about the local attractor. In Table \ref{tab:hybrid}, one can see that the hybrid PSO found the global optimum considerably more frequently than the social-only version, while its success rate is not too far behind the classical PSO. Additionally, as Table \ref{tab:hybrid} shows, the precision benefit of the hybrid PSO over the classical PSO was sometimes slightly worse than the benefit of the social-only PSO. The potential analysis of the hybrid PSO revealed a faster decrease of potential than in the case of the classical PSO. 
In Figure \ref{fig:rosenbrock_potential_hybrid}, one can see the course of the potentials from a sample run for both the classical and the hybrid PSO processing the $2$-dimensional \Rosenbrock function. The figure shows the potential curves for each algorithm and each of the two problem dimensions. Note that both runs used the same random seed, therefore the curves do not deviate before iteration $250$, when the local attractors of the hybrid PSO were disabled. After iteration $250$, one can clearly see a sharp bend in the curve, caused by the accelerated convergence of the hybrid
PSO that is now the social-only PSO.

\begin{table}
\caption{Number of results classified as `G' and their precision obtained by the classical PSO, the hybrid PSO and the social-only PSO processing various $3$-dimensional and a $4$-dimensional function. For \Griewank, \Rosenbrock and \Rastrigin (4-dim.), the hybrid PSO shows a significantly higher precision than the classical PSO. Additionally, for \Griewank, \Rastrigin and \Schwefel, it shows a clearly better exploration than the social-only PSO.}
\label{tab:hybrid}
\center{
\def\arraystretch{1.2}
\begin{tabularx}{0.95\textwidth}{|c| *{1}{|Y}|c| *{1}{|Y}|c| *{1}{|Y}|c|} 
\hline 

 \multicolumn{1}{ |c|| }{}
 & \multicolumn{2}{ c|| }{Classical PSO}
 & \multicolumn{2}{ c|| }{Hybrid PSO}
 & \multicolumn{2}{ c| }{Social-only PSO} \\
\hline 

 Function & G & Precision & G & Precision & G & Precision\\
\hline \hline

\Ackley & 50 & \hspace{0.1cm} 4.4409e-16 \hspace{0.1cm} & 50 & \hspace{0.1cm} 1.2257e-15 \hspace{0.1cm} & 50 & \hspace{0.1cm} 1.2967e-15 \hspace{0.1cm}\\
\Griewank & 25 & 3.6068e-12 & {\bf 19} & {\bf 0.0} & 2 & 0.0\\ 
\Elliptic & 50 & 4.0877e-37 & 50 & 3.6318e-38 & 50 & 8.2328e-40\\
\Rastrigin (3-dim.) & 50 & 0.0 & {\bf 50} & 0.0 & 28 & 0.0\\
\Rosenbrock & 47 & 0.0011 & 50 & {\bf 3.1236e-16} & 50 & 8.5184e-20\\
\Schwefel & 50 & 0.0776 & {\bf 50} & 0.0502 & 36 & 0.8312\\ 
\Sphere & 50 & 6.0717e-42 & 50 & 5.1757e-43 & 50 & 7.2010e-44\\
\hline \hline
\Rastrigin (4-dim.) & 50 & 4.6544e-07 & {\bf 47} & {\bf 0.0} & 10 & 0.0\\
\hline
\end{tabularx}

}
\end{table} 

 
\section{Concluding Remarks}

From our experiments, it is clearly evident
that the local attractor supports exploration and to a certain degree helps to avoid being trapped in a local optimum. The price of the local attractor is a sometimes reduced quality of the exploitation due to a slower convergence rate. 
For future work, one can form a more refined hybrid PSO variant between the social-only PSO and the classical PSO. Instead of simply switching from classical to social-only PSO after one half of the iterations, one could test different times for switching, or different forms of hybridization, e.\,g., a heterogeneous swarm with some particles that utilize their local attractor and other that do not.


\newpage

\bibliographystyle{plain}
\bibliography{literature}

\end{document}